\documentclass[a4paper,10pt,twoside]{article}

\usepackage[utf8]{inputenc}
\usepackage[T1]{fontenc}
\usepackage{palatino}

\usepackage[margin=1in]{geometry}
\usepackage{setspace}

\usepackage{amsmath}
\usepackage{amssymb}

\usepackage{graphicx}
\usepackage{grfext}
\usepackage{float}
\usepackage{longtable}
\usepackage{wrapfig}
\usepackage{rotating}
\usepackage{capt-of}
\usepackage{svg}

\PrependGraphicsExtensions*{.pdf,.PDF}

\usepackage[normalem]{ulem}

\usepackage{cuted}

\usepackage{tikz}
\usetikzlibrary{positioning,arrows.meta,backgrounds}

\usepackage{pgfplots}
\pgfplotsset{compat=1.18}

\pgfmathdeclarefunction{erf}{1}{%
  \pgfmathparse{2/sqrt(pi)*(#1-(#1^3)/3+(#1^5)/10)}%
}

\pgfdeclarelayer{bg}
\pgfsetlayers{bg,main}

\usepackage{minted}
\newminted{python}{fontsize=\footnotesize}

\DefineVerbatimEnvironment{verbatim}{Verbatim}{%
  fontsize=\scriptsize
}

\usepackage[
  backend=bibtex,
  style=numeric, 
  sorting=none,
  url=false
]{biblatex}
\usepackage{authblk}

\author[1]{Bartosz Barzdajn\thanks{bartosz.barzdajn@nask.pl}}
\author[1]{Mateusz Krzysztoń}
\affil[1]{NASK National Research Institute}

\title{Classical methods for modern applications: synthetic tabular data generation using lightweight discrete response continuous latent variable model.}
\date{\today}

\usepackage{hyperref}

\hypersetup{
  pdfauthor={Bartosz Barzdajn and Mateusz Krzysztoń},
  pdftitle={Classical methods for modern applications: synthetic data generation of discrete and continuous tabular data using lightweight continuous latent variable model.},
  pdfkeywords={machine learning, synthetic data, tabular data, latent variable model},
  pdfsubject={Synthetic tabular data generation},
  pdfcreator={LaTeX},
  pdflang={en}
}

\begin{document}

\maketitle
 \begin{abstract}
 \small

 \noindent
In this paper, we present a lightweight algorithm for modelling tabular data containing both continuous and discrete variables. The method uses empirical marginal distributions and a selected quantile function to transform the data into a continuous latent space, where a distribution of choice, e.g. a multivariate Gaussian, can be efficiently estimated and applied. While this approach is closely related to classical well-established machine learning methods, we present a novel adaptation that fits well within the contemporary computational ecosystem.

 \medskip

 \noindent
Unlike in the case of deep architectures, it results in a simple implementation with few dependencies, robust training/estimation, as well as efficient sampling. The transformation preserves complex marginal distributions and correlations, providing high-fidelity models, especially for data with unimodal distributions.

 \medskip

 \noindent
One of the key characteristics of the proposed approach is its continuous representation of discrete variables. While highly desirable, this introduces a significant challenge, since a single observation corresponds to a region rather than a point in the latent space. We show that this ambiguity naturally leads to a Monte Carlo expectation-maximisation (MCEM) formulation, whose auxiliary objective can be expressed as a weighted latent-space likelihood. This, in turn, allows conventional maximum-likelihood estimators to be used within an efficient and robust optimisation procedure.

 \end{abstract}
\section{Note on the future revisions of this draft}
\label{sec:org8956903}
While this manuscript presents the complete scientific contribution, future revisions may further elaborate on selected aspects, expand the discussion of related work, and adapt the presentation to the requirements of a specific journal. We greatly appreciate any comments, suggestions, or relevant references that could help us further improve the paper.
\section{Motivation}
\label{sec:org9222055}

The problem we address is the development of data-generating distributions for mixed-type data, with particular emphasis on the formal treatment of continuous and discrete features. This work is motivated by a cybersecurity use-case within the \href{https://www.mirandaproject.eu/concept/}{MIRANDA} project, where we are developing a fit-for-purpose predictive component for a cybersecurity digital twin (CDT).

The component will have two main tasks. The first is to generate synthetic data based on historical network traffic. These data can be used to test other CDT components, balance training sets, and pre-train more complex models while limiting overfitting. The second is to estimate the likelihood of observed network data, which may be applied to tasks such as evaluation of decision boundaries for anomaly detection.

The cybersecurity context imposes several practical requirements. The models must generate representative, high-fidelity samples. At the same time, it should be easy to implement with few dependencies, if any at all, allowing independent maintenance and limiting the attack surface. It also  must also be efficient enough to operate within a larger architecture with components competing for resources, and robust enough to be trained without extensive diagnostics or monitoring. As a result, rather than chasing "state-of-the-art" we are aiming at "fit-for-purpose".

These requirements led us to revisit classical methods rather than rely on deep architectures. In particular, we draw inspiration from both classical discrete-response latent variable models \autocite{Muthen1984} and data transformations closely related to Gaussian copula modelling \autocite{Sklar1973}. As we will see, this initially proved to be a difficult marriage, ultimately leading to a novel implementation of the expectation-maximisation algorithm. The resulting method is surprisingly simple to implement despite its formal treatment of mixed-type data and strict handling of discrete variables. We believe that the derivation of this algorithm, and its implementation, may have broader relevance, especially in contexts with requirements similar to those described above. One of the advantages of our approach is its simple, yet explicit, continuous latent representation for all feature types, which opens a path for straightforward interpolation and extrapolation of mixed-type data-generating distributions.
\section{The core concept and its relation to other methods}
\label{sec:org9ec34b4}

Before moving forward, it is worth emphasising that the literature on generative modelling, latent-variable methods, and copula-based approaches is extensive. Here, we focus on a small number of representative methods that, in our view, help place the proposed approach in a broader context and highlight its relationship to contemporary methods, while at the same time providing a smooth introduction to the rather involved derivation that follows.

In many real-world applications, training data follow complex distributions that are difficult to model directly. The core idea behind many solutions is to transform the data into a space where much simpler, well-established distributions can be applied. This approach underlies methods such as normalising flows \autocite{Rezende2015}, VAEs \autocite{Kingma2013} and even, in some sense, GANs \autocite{Goodfellow2014}. The main differences lie in how the transformation is structured and learned.

While in principle these transformations may be arbitrary, as argued above, we favour robust statistical methods that can be evaluated directly from the data. Furthermore, since we wish to perform both generation and inference, we require explicit access to the latent space, just as in the case of VAEs. This allows likelihoods to be evaluated using the latent distribution (see e.g. \autocite{Prince2024} or Appendix \ref{sec:org0501fd4}).

Therefore, we choose a simple two-stage transformation, similar to that used in Gaussian copula modelling, although we do not restrict ourselves to the specific latent distribution typically associated with copula models. In principle, the second stage may be defined by any continuous distribution with a well-defined quantile function. Each feature is first mapped through its empirical CDF to the quantile space and then through a selected quantile function, which defines the desired marginal distribution in the latent space.

\begin{figure}[H]
\centering
\includegraphics[width=1.0\textwidth]{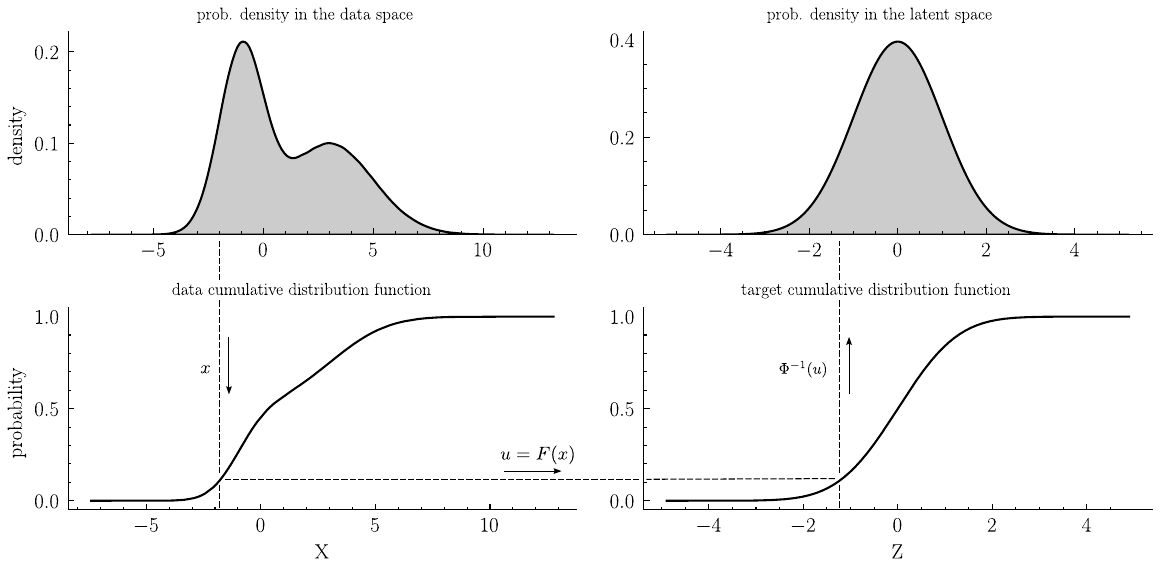}
\caption{\label{fig:orgd4e0139}Illustration of the two-stage quantile transformation used to construct the latent representation. While the transformation reshapes the marginals into the desired form, the dependence structure remains non-trivial.}
\end{figure}

This is a well-established approach originating from Gaussian copula modelling, albeit somewhat different from the way copulas are commonly introduced. While the copula itself is formally defined in the quantile space, practical Gaussian copula models are often estimated through observations transformed into a latent Gaussian representation. The same two-stage transformation serves as the starting point for our approach. Figure \ref{fig:orgd4e0139} illustrates the transformation for a single feature and how it is mapped to a Gaussian target distribution.

Note that this operation is applied independently to each feature and therefore only reshapes the marginal distributions into the desired form. For multivariate data, much of the complexity of the distribution may still be preserved through the correlation structure. This is something we must keep in mind when constructing the model of our data.

While this approach seems straightforward, in our context we need to address another layer of complexity, namely the treatment of discrete variables. Real-world network data usually contain both continuous and discrete features, and this remains a recognised challenge for generative modelling frameworks.

One solution, often used in normalising flows, is dequantisation (see, e.g. \autocite{Ho2019}). In other words, replacing discrete variables with their continuous surrogates. In practice, this means adding noise to discrete observations before training, so that they can be treated as continuous. From the point of view of marginal transformations, this implicitly replaces a step-wise empirical CDF with a smoothed approximation, where the smoothing kernel depends on the chosen noise distribution.

Copula Flows represent a related but more formal route \autocite{Kamthe2021}. Since discrete CDFs are step-wise and copulas are not unique for discrete data, the method uses the distributional transform \autocite{Ruschendorf2009}. A discrete observation is mapped stochastically to a value inside the corresponding jump of the CDF, which gives a continuous uniform representation suitable for flow-based copula modelling. From our perspective, this is particularly interesting because a single discrete observation is no longer associated with a single quantile value, but with a whole range of admissible quantile representations. This will become one of the core concepts behind our method.

Another approach can be found in CTGAN \autocite{Xu2019a}, which is often used as a default method for synthetic tabular data generation. CTGAN handles mixed-type tabular data by changing the representation of both variable types before training. Discrete features are represented by one-hot vectors, while continuous features are represented by two quantities: a mode indicator of a pre-fitted Gaussian mixture and the value normalised within that mode. In other words, CTGAN introduces, in some sense, an intermediate representation, in which continuous variables are partly described through discrete mode labels.

Naturally, as mentioned before, such a review can never be even close to complete. While many recent developments focus on achieving state-of-the-art performance, often through diffusion models \autocite{Kim2023,Kotelnikov2022} or based on foundation models for tabular data such as TabPFN \autocite{Hollmann2023}, some researchers, similarly to our work, favour classical statistical modelling over deep architectures. One such example is Synthsonic \autocite{Baak22}, which combines a sequence of statistical transformations with Bayesian network modelling and discriminate learning. 

Continuing the line of thought introduced by the distributional transform, we take a formal approach to the problem of mixed-type features, originally inspired by the ClustMD model \autocite{Parland2016}, but in fact rooted more broadly in the discrete-response latent variable framework introduced by Muthén \autocite{Muthen1984}.

The latent representation remains continuous, while discrete observations correspond to regions of the latent space. Unlike dequantisation, we do not replace discrete observations with continuous surrogates, but instead model the ambiguity they imply.

The price we pay for this formal treatment is that inference becomes more involved. As we shall see, the resulting optimisation problem naturally leads to a Monte Carlo expectation-maximisation (MCEM) formulation. However, the corresponding auxiliary objective can be expressed as a weighted likelihood in the latent space, turning what initially appears to be a difficult inference problem into a fairly straightforward optimisation procedure.

As indicated before, the benefit is that the latent representation remains continuous for all feature types. This makes the model easy to estimate, efficient to sample from, and robust when interpolating between observed conditions. It also provides a natural framework for future regression models, where changes in device or deployment characteristics can be expressed through changes in the parameters of the latent distribution.
\section{The method}
\label{sec:orgf156d21}

This section is intended as a high-level overview of the method, while a more complete treatment can be found in Appendix \ref{sec:org5322f6d}. We aim to keep the discussion accessible to readers who may be less familiar with classical machine learning and statistical modelling. However, for the sake of brevity and clarity, we intentionally omit some technical subtleties and focus more on the essential concepts and the novel elements of the derivation. Readers primarily interested in the utility of the method may refer to the following section.

The starting point for our reasoning is the Evidence Lower Bound (ELBO) \autocite{Murphy2013}, \autocite{Bishop2006} -- the go-to identity for latent variable modelling. The goal is to approximate the solution that maximises the likelihood of data under specific values of the model parameters, while the model is defined on the latent space, hidden behind a reversible transformation, where latent states are not directly observable.

In general, the core idea is to "expose" the latent variables and define the joint distribution together with ones that are observable. Then, instead of dealing with intractable integration's that arise, we find the lower bound of the likelihood. By maximising this bound, we match a surrogate objective for likelihood maximisation. For convenience, the complete framework is provided in the Appendix \ref{sec:org4313ea1}.

We begin by exposing the latent state through a marginalised joint distribution. For now, we do this for a single point \(x\) in the data space, writing the likelihood as
\begin{equation}
\label{eq:org57a3132}
p(X=x \mid \theta) = \int_{\mathcal{Z}} p(X=x, Z=z \mid \theta)\, dz,
\end{equation}
where integration is performed over the volume \(\mathcal{Z}\) of latent states corresponding to the observation \(x\). In the above, the likelihood depends on the parameter vector \(\theta\), \(X\) is the random variable representing the data, and \(Z\) corresponds to the latent states. Here, we use a simplified notation where \(x\) and \(z\) are vectors in a multidimensional real space.

In the continuous case, we could follow the standard ELBO framework, define the corresponding expectation-maximisation objective, and proceed with the optimisation (again, see e.g. \autocite{Murphy2013}, \autocite{Bishop2006} or Appendix \ref{sec:org4313ea1}). In our case, however, the transformation to the latent space is no longer one-to-one, as the model assumes a discretised response. We therefore need to consider several key nuances.

Formally, we can define the forward transformation, an "encoder", as 
\begin{equation}
x_{ik} \rightarrow u_{ik} = \widehat{F}_k(x_i) \rightarrow z_{ik} = \Phi^{-1}_k(u_i)
\end{equation}
where \(\hat{F}_k\) is the marginal empirical cumulative distribution function (CDF) of data. For the \$k\$-th feature, the first step applies column-wise the standard quantile transformation, independently with respect to each feature, while the second applies another per-feature mapping through the quantile function of the selected one-dimensional distribution. This is the same procedure as the one illustrated in Figure \ref{fig:orgd4e0139}. In the above, \(\hat{F}\) and \(\Phi\) are vector-valued functions with independent components. We use the standard normal distribution as the target distribution, hence the notation \(\Phi\). In principle, this target could be replaced by any other proper distribution, although we will see later that this choice plays an important role.

Let us focus now on the key consequence of the adopted transformation. In the discrete-response case, a single, i-th observation does not identify a unique latent value. Instead, together with the "decoder"" \(f\), it defines a per-data-point pre-image 
\begin{equation}
\mathcal{Z}_i = \left\{ z : f(z) = x_i \right\},
\end{equation}
i.e. a set of points corresponding to the specific \(x_i\). Since the latent space is continuous, \(\mathcal{Z}_i\) represents a volume. This will become important later in defining explicitly our joint distribution of latent states and data, as well as the associated probabilistic measure.

As indicated before, in a regular variational framework, where we take advantage of the ELBO, we do not optimise the likelihood or log-likelihood directly, as in standard maximum likelihood estimation. Instead, we aim to maximise the lower bound of equation \ref{eq:org57a3132}. Going one step further, it is sufficient to maximise an auxiliary function that consists only of terms depending on the parameters of the latent distribution. In a similar fashion, we can define the auxiliary function \(A(\theta)\) that will serve as our optimisation objective. After dropping the entropy term from the lower bound, it takes the following general form:
\begin{equation}
A(\theta) = \sum_{i} \int_{\mathcal{Z}_i} \log p(X=x_i, Z=z \mid \theta) \, q_i(z) \, dz.
\end{equation}
The result above is very similar to the standard variational framework. The key difference lies in the limits of integration, which arise from restricting \(z \mid x_i\) to the admissible region \(\mathcal{Z}_i\). This restriction is a direct consequence of the relationship between the step-wise transformation and the joint distribution, which we will address later. It also induces an observation-specific variational posteriors \(q_i\), supported only on the corresponding pre-image.

Although  \(q\)'s may be arbitrary, we can show that in a more conventional setting the optimal choice, one that maximises the lower bound, is the one that also minimises the KL divergence - see Appendix \ref{sec:orgb690d9a},
$$
D_{\mathrm{KL}}\!\left(q(z) \,\Vert\, p(z \mid x,\theta)\right).
$$
As \(q\) approaches the latent posterior \(p\), the bound becomes "tight" and we can approximate our objective.

As a result, we can define variational posteriors \(q_i\) as
\begin{equation}
q_i(z)=p(Z=z \mid X=x_i,\theta).
\end{equation}
The notation here is almost identical to the conventional latent posterior on purpose, as we will be using single latent "parent" model for all observations. However, each time the support will depend explicitly on the observed response and each one of them will correspond to different restrictions, resulting in different normalisation. This observation-specific posterior will play the role of the probabilistic measure and define the expectation in the optimisation objective.

Now we can write-down the near-final form of the objective. After applying the probabilistic chain rule, to factorise the joint distribution, at the end we obtain
\begin{equation}
\def\J{p(X=x_i \mid Z=z, \theta) p(Z=z \mid \theta)}
\def\P{p(Z=z \mid X=x_i, \theta)}
A(\theta) = \sum_{i} \int_{\mathcal{Z}_i} \log \left( \underbrace{\J}_{\text{joint distribution of data and latent var.}}  \right) \underbrace{\P}_{\text{probabilistic measure }} \, dz.
\end{equation}
At this stage the structure of the objective becomes transparent. As we "go" over the latent space, the first factor in the joint distribution, \(p(X = x_i \mid Z = z, \theta)\), is degenerate and equal to one when \(z\) is a part of the pre-image, and zero elsewhere. Hence, we introduced earlier the domain of integration \(\mathcal{Z}_i\). By restricting it, contributions of the first factor become constant and it can be ignored.

To summarise, the probability measure is obtained by restricting the same parent latent model to the corresponding pre-image and renormalising. In other words, all observations share a common latent model, but each observation only "sees" the part of that model that is compatible with its own discrete response. This is precisely the mechanism that allows us to combine a continuous latent space with discrete observations.

The remaining difficulty lies in distributions \(p(Z = z \mid X = x_i, \theta)\), which defines how probability is distributed over each \(\mathcal{Z}_i\). Although all \(z \in \mathcal{Z}_i\) are consistent with the observation \(x_i\), they are not equally probable under the model. Their likelihood depends on the latent prior \(p(Z=z \mid \theta)\), which in turn corresponds to the model we wish to find.

This circular dependence naturally leads to an expectation-maximisation (EM) formulation. Given the current parameter estimate \(\theta^{(k)}\), we first evaluate the posterior distribution over the latent variables restricted to \(\mathcal{Z}_i\), namely
\(p(Z=z \mid X=x_i,\theta^{(k)})\).
This distribution determines the relative density of the admissible latent states and defines the expectation step. We then update \(\theta\) by maximising the expected log-likelihood of the remaining factor in the joint distribution, \(p(Z=z \mid \theta)\), with the expectation taken under the posterior evaluated at \(\theta^{(k)}\). This defines the maximisation step.

The computational challenge here is the calculation of integrals for each element of the training set. This leads to one of the key observations behind our approach. The objective can be written as a sum of expectations with respect to the measures \(p(Z = z \mid X = x_i, \theta)\), and such expectations can be approximated using the Monte Carlo (MC) method. In a generic form, this approximation is given by
\begin{equation}
\int_\mathcal{X} h(x) \, p(x)dx = \mathbb{E}_p[h(X)] \approx \frac{1}{N}\sum_{i=1}^N h(x_i).
\end{equation}
This simplifies the problem significantly, since we can optimise a sum of averages instead of a sum of integrals. However, this procedure is straightforward only if \(x_i \sim p\), that is, if the samples are distributed according to \(p\). Here, we are dealing with a family of truncated distributions, each defined over its own truncation region, from which efficient direct sampling may be difficult. In principle, we could use more advanced methods, such as Markov chain Monte Carlo (MCMC). In practice, however, applying such methods to an ensemble containing potentially as many distinct distributions as there are unique data points would be difficult to implement and computationally expensive. 

\begin{figure}[H]
\centering
\includegraphics[width=1.0\textwidth]{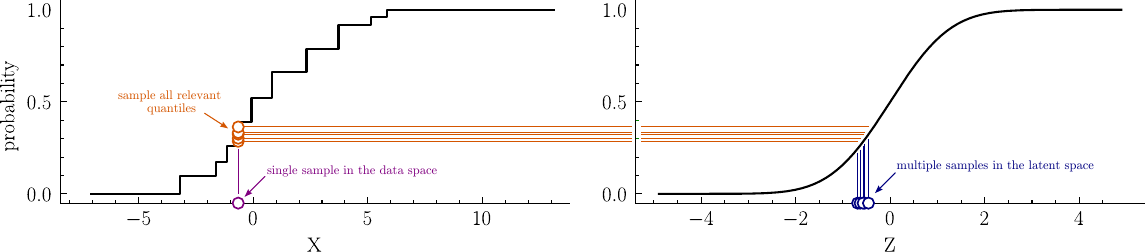}
\caption{\label{fig:orgf80f92a}Observation-specific sampling in the continuous latent space. Each discrete observation defines an admissible region \(\mathcal{Z}_i\), from which multiple latent representations are generated for Monte Carlo approximation.}
\end{figure}

Instead, we can use self-normalized importance sampling (SNIS). Let us focus first on the sampling procedure. When we have a discrete observation it corresponds not to a single quantile value, but to a range of possible quantile values defined by the step-wise marginal empirical CDF. We sample these values, transform them into the latent space, and use appropriate importance weights to estimate the required integrals. This provides a good coverage of the latent space and corresponding pre-image, since each data point is represented by controllable number of latent samples. As such, it also forms the basis for the efficient computation of multiple expectations/integrals. The concept of sampling illustrated in Figure \ref{fig:orgf80f92a}.

The proposal distribution will have independent marginals, i.e.
\begin{equation}
q(x_1,\ldots,x_d) = q_1(x_1)q_2(x_2)\cdots q_d(x_d) = \prod_{k=1}^{d} q_k(x_k).
\end{equation}
This way, not only can we sample each feature separately, but the calculation of the proposal density is also very simple.

In general, the importance sampling estimate can be written as
\begin{equation}
\mathbb{E}_{p}[h(X)]
\approx
\frac{\sum\limits_{i=1}^N w_i h(X_i)}{\sum\limits_{i=1}^N w_i},
\end{equation}
where
\begin{equation}
w_i := \frac{\phi(X_i)}{q(X_i)}.
\end{equation}
Here, \(\phi\) is the target distribution, that is, our parent distribution, for which we wish to calculate the expectation, and \(q\) is a for now described proposal distribution from which we can actually sample. We can now see why the choice of the distribution used as the basis of our transformation is so important. It defines the proposal distribution. We can also see, that weights are by definition self-normalising. Hence the name self-normalized importance sampling (SNIS).

It remains to specify the EM auxiliary function that defines the M-step objective. At iteration \(k\), for a single observation, we approximate its contribution by
\begin{equation}
A_i \left( \theta, \theta^{(k)} \right) = \int_{\mathcal{Z}_i} \log p \left( X=x_i \mid Z=z, \theta \right) p \left( Z=z \mid \theta \right) \, p \left( Z=z \mid X=x_i, \theta^{(k)} \right) \, dz,
\end{equation}
where the first term is constant within the corresponding integration volume. Approximating the integral with \(M\) samples from the pre-image \(\mathcal{Z}_i\) gives the following self-normalized importance sampling estimate:
\begin{equation}
\def\W{\frac{p \left( Z=z_{ij} \mid X=x_i, \theta^{(k)} \right)}{\phi(z_{ij})} \frac{1}{\sum\limits_{j=1}^M \frac{p \left( Z=z_{ij} \mid X=x_i, \theta^{(k)} \right)}{\phi(z_{ij})}}}
A_i(\theta, \theta^{(k)}) \approx \sum_{j=1}^M \log p \left( Z=z_{ij} \mid \theta \right) \underbrace{\W}_{\text{normalised weight } \tilde{w}_{ij}},
\end{equation}
where \(z_{ij} \in \mathcal{Z}_i\) are samples from the pre-image of observation \(x_i\), \(\phi(z)\) is the distribution used in the second stage of the transformation and serves as the proposal distribution, for example the standard normal distribution, and \(p(Z=z_{ij} \mid X=x_i, \theta^{(k)})\) is the target sampling distribution, which depends on the parameter estimate from the previous optimisation step. The last terms is the normalisation factor -- sum over all "weights".

Summing the individual contributions over the training set gives the complete auxiliary objective:
\begin{equation}
A(\theta, \theta^{(k)}) \approx \sum_{i=1}^N \sum_{j=1}^M \tilde{w}_{ij} \log p \left( Z=z_{ij} \mid \theta \right),
\end{equation}
where \(\tilde{w}_{ij}\), just as indicated in the previous equations, are normalised, per-observation, weights. The maximization step therefore reduces to weighted maximum likelihood estimation of the latent model. For the multivariate Gaussian considered here, this means estimating the mean vector and covariance matrix from weighted samples, where the weights are determined by the current estimate of the model parameters.
\section{The final algorithm}
\label{sec:orgba64f63}
The model parameters are estimated iteratively using the EM algorithm, which typically converges within 10--30 iterations. At each iteration, the E-step produces a weighted sample representation of the latent space, reducing the M-step to a standard weighted maximum-likelihood problem.

For the Gaussian latent prior with parameters \(\theta = (\mu, \mathbf{C})\), the updated mean is given by
\begin{equation}
\mu = \sum_{i=1}^N \sum_{j=1}^M \tilde w_{ij} z_{ij},
\end{equation}
where \(z_{ij}\) denotes a sample drawn from the proposal distribution \(\phi\) over the pre-image \(\mathcal{Z}_i\) associated with observation \(x_i\). The normalised importance weights are
\begin{equation}
\tilde w_{ij}
=
\frac{w_{ij}}
     {\displaystyle\sum_j w_{ij}},
\end{equation}
with unnormalised weights
\begin{equation}
w_{ij}
=
\frac{p \left( Z=z_{ij} \mid x_i, \theta^{\mathrm{(prev)}} \right)}{\phi(z_{ij})},
\end{equation}
where \(p(\cdot \mid x_i, \theta^{\mathrm{(prev)}})\) is in principle the observation-specific posterior induced by the latent prior from the previous iteration. Thanks to the use of self-normalised importance sampling (SNIS), and because the observation-specific posteriors differ only by a normalising constant, the importance weights can be evaluated using the underlying "parent" latent model. In our case it would be the multivariate Gaussian distribution. 

Finally, using the biased covariance estimator, the covariance matrix is updated as

\begin{equation}
\mathbf{C}
=
\sum_{i=1}^N \sum_{j=1}^M
\tilde w_{ij}
z_{ij} z_{ij}^{\top}.
\end{equation}
In practice, we use an unbiased covariance estimator, which requires a slightly modified update. Regardless of the specific estimator, each EM iteration follows the same principle: the current prior parameters define the importance weights, which are then used to estimate an updated set of latent model parameters.

It immediately becomes apparent that, despite the rather involved derivation, the resulting algorithm is remarkably simple. Most of the implementation complexity stems from managing the sampling procedure and normalising estimates obtained from observation-specific regions of the latent space.
\section{Performance evaluation}
\label{sec:org89c7f62}

We decided that the best way to evaluate the utility of our model is through a series of controlled experiments using synthetic data drawn from a non-trivial parametric distribution with a controllable level of difficulty. For each difficulty level, we sample a training set, fit the evaluated models, we will explain the plural in a moment, and compare the data they generate with an independent reference sample from the same distribution. We repeat this procedure across different parameters, observe how performance changes and possible identify the conditions under which our model remains useful. To provide meaningful points of comparison, we evaluate methods of varying complexity, including a zero-shot baseline, our model, and an industry-standard CTGAN.

The benchmark distribution should reflect the main challenges posed by the datasets addressed in this work. These include mixed data types, both discrete and continuous, strong correlations, and occasional near-deterministic relationships between features. Such relationships may arise from feature engineering or when the same quantity is obtained from different sources, for example, when one value is inferred and another is measured directly. The benchmark should also expose a key limitation of our model: its difficulty in representing distributions with ambiguous marginals, such as the commonly used two-moons benchmark or distributions with a ring-like topology. We therefore require a parametric family that can vary between these challenging structures and simpler distributions that our model can easily represent.

\begin{figure}[H]
\centering
\includegraphics[width=1.0\textwidth]{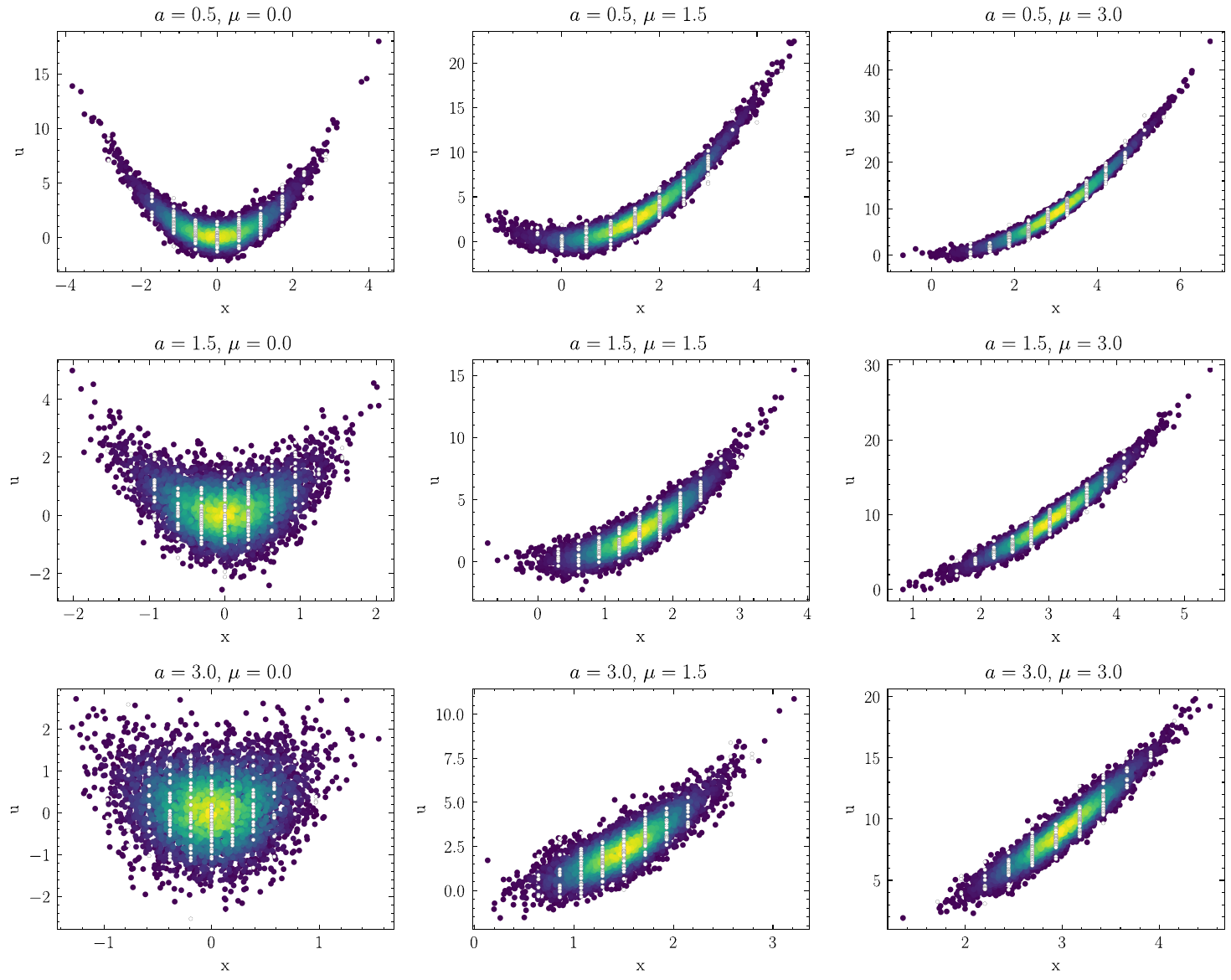}
\caption{\label{fig:org5f29df0}Unnormalised densities of the Rosenbrock distribution and corresponding discretised samples for selected parameter settings.}
\end{figure}

We found that the Rosenbrock distribution meets these requirements particularly well. It is easy to sample from, and its unnormalised density can be evaluated directly. Despite having only two parameters, it can vary from an approximately elliptical distribution to one with strongly ambiguous marginals. To reflect the mixed-type setting central to this work, we additionally discretise selected dimensions of the sampled data. We decided that two-dimensional version will serve us well. Figure \ref{fig:org5f29df0} illustrates examples of distributions for selected parameters and respective discretised samples.

\begin{figure}[H]
\centering
\includegraphics[width=1.0\textwidth]{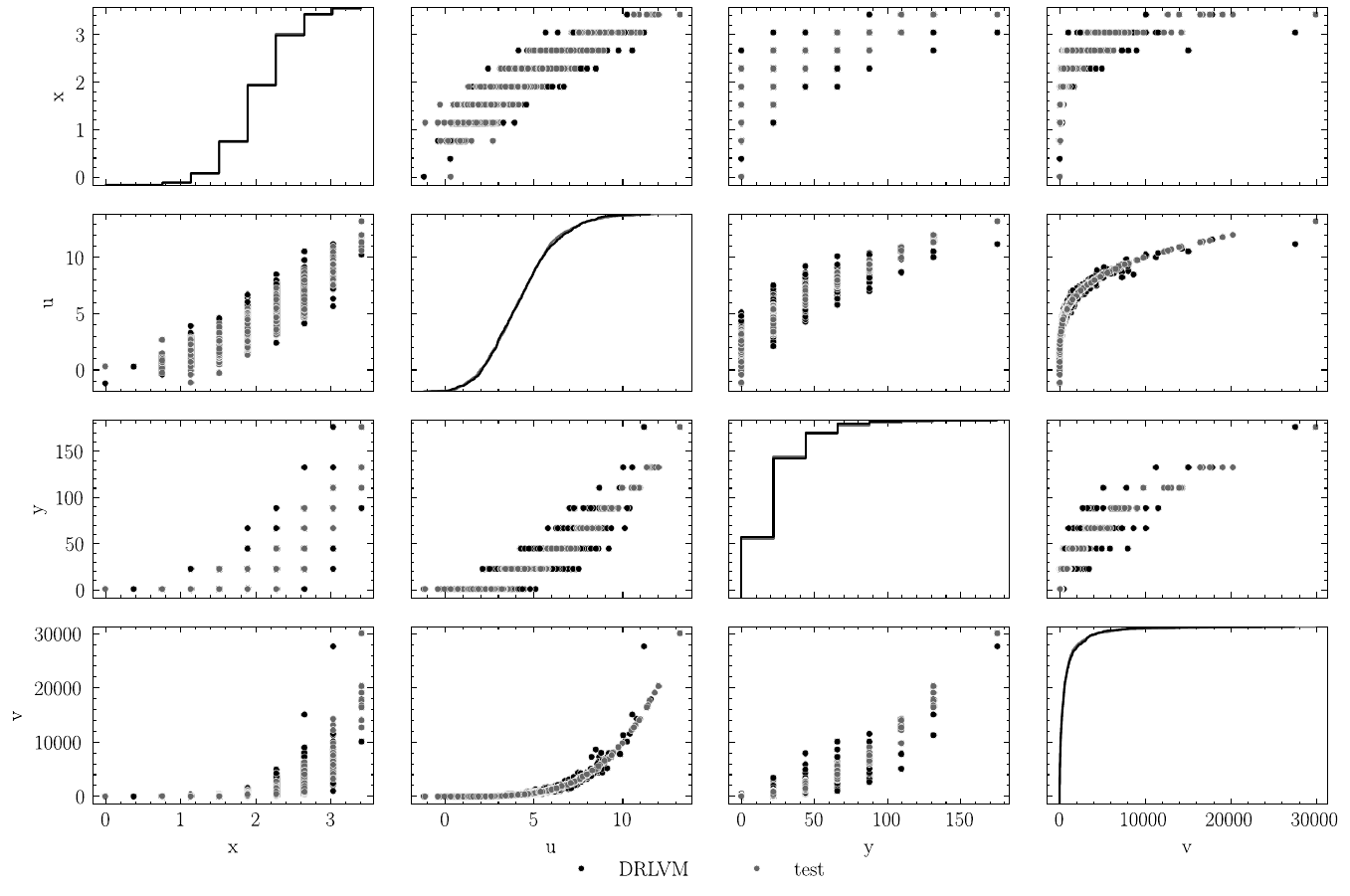}
\caption{\label{fig:orgf4544e0}Reference test data (gray) and sample obtained using proposed method - DRLVM (black). Results represent Rosenbrock distribution case study with scale and location parameter \(a = 2.5\) and \(\mu = 2.0\) repectively.}
\end{figure}

The case study covers 35 four-dimensional Rosenbrock distributions, obtained from all combinations of $a \in \left\{ 0.5, 1.5, 2.0, 2.5, 3.0 \right\}$ and $\mu \in \left\{ 0.0, 0.5, 1.0, 1.5, 2.0, 2.5, 3.0 \right\}$. For each pair of parameters, we draw 12,500 observations and discretise two dimensions, $x$ and $y$, while leaving $u$ and $v$ continuous. Figure \ref{fig:orgf4544e0} illustrates a comparison between the reference (test data) and data generated using proposed method. This example may serve well as a qualitative easement of a example with intermediate difficulty.

\begin{figure}[H]
\centering
\includegraphics[width=1.0\textwidth]{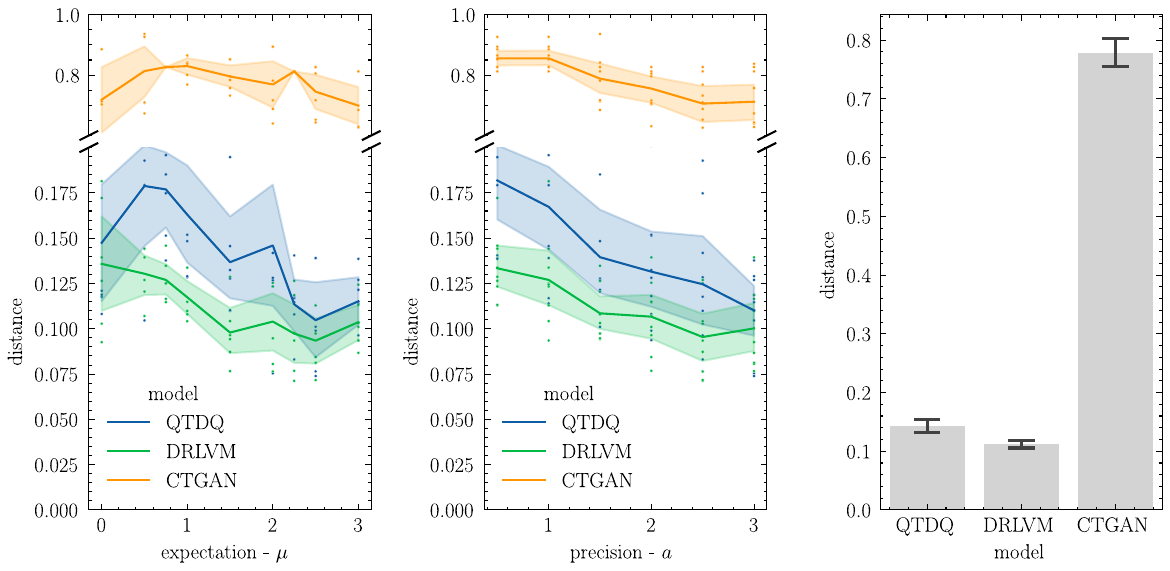}
\caption{\label{fig:orgb27a3cc}Energy distance between the generated and reference samples for the QTDQ baseline, DRLVM, and CTGAN. The left and centre panels show the distance as a function of the Rosenbrock location parameter \(\mu\) and precision parameter \(a\), respectively. Lines show the mean across the remaining parameter settings, while shaded regions indicate the corresponding 0.95 confidence interval. The right panel summarises the results across all 35 benchmark scenarios. Error bars indicate the variation between scenarios. Lower values indicate better agreement with the reference distribution.}
\end{figure}

In the quantitative assessment, we divide the resulting mixed-type dataset into training and test sets using an 80:20 split. We evaluate three methods of increasing complexity: a quantile-transformation baseline with a fitted Gaussian distribution\footnote{This method consist of dequantisation by adding Gaussian noise, transforming data into quantile space by applying feature-wise empirical CDF, projecting each column into one dimensional Gaussian quantile function and fitting a Multivariate normal distribution. During generation discrete components are snapped to the closest value in the training set. This method can be viewed as dequantised version of Gaussian copula modelling and a one-shot version of the proposed DRLVM.}, the proposed DRLVM, and CTGAN with optimised hyperparameters. We use each fitted model to obtain 2500 observations, which are compared with the corresponding test set using energy distance,
\begin{equation}
\widehat{D}(X, Y)
=
\frac{2}{nm}\sum_{i=1}^{n}\sum_{j=1}^{m}
\left\lVert x_i-y_j\right\rVert_2
-
\frac{1}{n^2}\sum_{i=1}^{n}\sum_{j=1}^{n}
\left\lVert x_i-x_j\right\rVert_2
-
\frac{1}{m^2}\sum_{i=1}^{m}\sum_{j=1}^{m}
\left\lVert y_i-y_j\right\rVert_2,
\label{eq:empirical_energy_distance}
\end{equation}
where \(X = \{x_i\}_{i = 1}^{n}\) denotes the generated sample, \(Y=\{y_j\}_{j=1}^{m}\) denotes the corresponding test sample, and \(\lVert \cdot \rVert_2\) is the Euclidean norm. Lower values indicate greater agreement between the generated and reference distributions. Results this comparison are presented in figure \ref{fig:orgb27a3cc}. The results, including their surprising aspects will be discussed in the following section.

\section{Summary, discussion and future work}
\label{sec:orgbf4bc9c}

Taking advantage of quantile transformations can provide robust, yet high-fidelity statistical models of data with simple latent priors. In this context, the formal treatment of discrete responses leads to a rather involved derivation, but the resulting algorithm can be remarkably simple. With the right approach, each EM iteration was reduced to evaluation of importance weights and performing simple weighted maximum-likelihood estimation. For the Gaussian latent model considered here, this further reduces to calculating a weighted mean and covariance matrix. The introduced method is therefore computationally efficient and straightforward to implement, reducing resource usage, the number of required external dependencies and the attack surface as a result.

Synthetic tests confirmed that the method is capable of generating high-fidelity samples, even for difficult data distributions, despite the fairly simple underlying latent model, which turn is also a significant advantage. Furthermore, comparison with its one-shot counterpart, demonstrated that in many cases, there is measurable benefit from applying a more formal approach rather then relying on dequantisation. Finally, explicit access to the latent space allows us to evaluate the likelihood of samples in the data space, providing a direct way to obtain, for example, class probabilities or probabilistic decision boundaries in anomaly detection.

The comparison with CTGAN produced an unexpected result. Although CTGAN can represent a much broader family of distributions and reproduced the overall shape of the Rosenbrock distribution more accurately, it achieved substantially higher energy distances than the proposed method. A closer examination suggests that CTGAN failed to match the narrow concentration of probability mass around specific dimensions of the data manifold, despite the use of optimised hyperparameters. More detailed results and visual comparisons will be provided in the accompanying computational notebook that will be referenced in the next revision of this draft.

While certainly CTGAN is capable of obtaining much better results, for example with tailored preprocessing, after all, our model most likely is a subset of distributions that CTGAN can in principle represent, the result demonstrates how important is robustness of ML solutions. This aspect is quite often neglected and lost in sophisticated pipelines.

When it comes to the future of our approach we would like to emphasise the following. The introduced methodology consists of replaceable stages, each with a well-defined purpose. Both the transformation and the latent model may be substituted without changing the overall structure of the algorithm. In principle, any model with a practical weighted-likelihood estimator can be "inserted" into the EM loop, and any distribution with a well-defined quantile function can be used in the second stage of the transformation. It will be interesting to see whether richer latent models, such as mixture models, capable of representing a much broader family of data distributions, will maintain desired "simplicity" of the implementation.

Finally, simple parametric latent models offer a natural extension towards conditional generation, including interpolation and prediction. An unconditional empirical CDF may, however, remove trends that the regression model is intended to represent. On the other hand, it may be possible to replace it with a conditional transformation based on quantile regression. The resulting conditional "decoder" could then describe how entire marginal distributions change with the explanatory variables. Although, this would address only part of the problem, since the dependence structure may also change. Somewhat in the spirit of copula modelling, one could therefore model the conditional marginals and the parameters of the latent dependence structure separately. Such an extension appears natural within the proposed framework, although its estimation is likely to be considerably more involved.
\section{Acknowledgements}
\label{sec:org89f507f}

This paper was written as part of the MIRANDA project that received funding from the European Union’s Horizon Europe Research and Innovation Programme under grant agreement No. 101168144 (MIRANDA). The views and opinions expressed are those of the authors only and do not necessarily reflect those of the European Union or the European Cybersecurity Competence Centre. Neither the European Union nor the granting authority can be held responsible.
\section{References}
\label{sec:orgfae4f93}

\printbibliography[heading=none]

 \appendix
\section{Complete derivation of the DRLVM model}
\label{sec:org5322f6d}
\subsection{Initial definition of the optimisation objective}
\label{sec:orgbfa6f14}

The goal is to model the data-generating distribution efficiently. As mentioned before, we achieve this by introducing a transformation to a latent space, together with a well-defined inverse process, such that the distribution is simpler in the latent space. Generating new synthetic data then involves sampling from this space and transforming the samples back to the original data space. This approach can maintain high fidelity while relying on relatively simple parametric distributions, but it requires us to model unobserved latent states. A common solution in such cases is to use variational methods. Therefore, our derivation begins with one of the key identities of latent-variable modelling, the Evidence Lower Bound (ELBO).

We encourage readers to visit other resources for further details. For convenience, we present our own derivation in Appendix \ref{sec:org4313ea1}. Note that, while our method is rooted in this framework, for the sake of clarity, we will only provide a brief overview and focus on the key, from our perspective, elements.

Our goal is to find a latent distribution that maximises the likelihood of data. The core idea is to "expose" the latent variables and define the a joint distribution together with variables that are observable. We do this by introducing its marginalised ("integrated-out") version
\begin{equation}
p(X=x \mid \theta) \;=\; \int_{\mathcal{Z}} p(X=x, Z=z \mid \theta)\, dz.
\end{equation}
In the above,  left -hand side is the likelihood of data given the parameters \(\theta\), \(X\) is the random variable representing the data and \(Z\) corresponds to relevant latent states.

From the multiplicativity of probability and the fact that measurements themselves are independently distributed, we know that it is preferable to work with log-likelihoods. However, this standard transformation, that usually allows us to simplify the problem straightaway, because of the integral does not gives us much in this case. 

Instead of maximising the likelihood directly, after rather lengthy transformations that involve Jensen's inequality, we can derive the lower bound on the log-likelihood (see Appendix \ref{sec:org4313ea1}). This leads to a version of the problem that is much easier to work with. For a single data point it is defined as
\begin{equation}
\ell(\theta) \geq \mathbb{E}_{q(z)}[\log p(X=x_i, Z=z)] + H(q(z)).
\end{equation}
where \(\theta\) represents the model parameters, \(x_i\) is an individual observed data point. The term \(q(z)\) is the variational distribution, which in principle is arbitrary, but in fact, as we will see later, should approximate the true posterior. Note that \(\mathbb{E}_{q(z)}[\cdot]\) denotes the expectation taken with respect to this distribution. Informally
\begin{equation}
\label{eq:orgc1f24ce}
\mathbb{E}_{q(z)} \left[ \ldots \right] = \int \, \ldots \, q(z) dz.
\end{equation}
Finally, \(H(q(z))\) is the Shannon entropy of the variational distribution, defined as \(H(q(z)) = -\mathbb{E}_{q(z)}[\log q(z)]\), which regularises \(q\) and prevents it from "collapsing". By maximising the lower bound with respect to the model parameters, we are also approaching the maximum-likelihood solution. Unfortunately, as a result of this transformation, we have to deal with the ambiguity introduced by \(q(z)\).
\subsection{The transformation}
\label{sec:org845aff1}

As mentioned earlier, this model constructs a joint distribution that involves a mapping through the empirical cumulative distribution function (CDF). The transformation proceeds in two steps. First, an observation \(x_{i}\) is mapped to the quantile space:
\begin{equation}
u_{ij} = \hat{F}_j(x_{ij}) = \frac{1}{N} \sum_{i\prime}^N \mathbb{I}(x_{i\prime j} \leq x_{ij}),
\end{equation}
and subsequently expressed as a latent variable:
\begin{equation}
z_{ij} = f(x_{ij}) = \Phi^{-1}_j(u_{ij}) = \Phi^{-1}_j\bigl(\hat{F}_k(x_{ij})\bigr).
\end{equation}
Here, \(\mathbb{I}(\cdot)\) denotes the indicator function and \(\hat{F}\) is the empirical CDF. Throughout this work we use the standard normal distribution as the target distribution, hence the use of the symbol \(\Phi\) to denote its CDF. Note that all transformations are performed independently for each column \$j\$-th (feature), and therefore \(\hat{F}\) represents the marginal CDF for this specific feature.

This structure combines the empirical distribution with a Gaussian target, creating, in the spirit of a Gaussian copula, a "normalised" representation of the data. The inverse mapping proceeds by first returning to the quantile space via \(u_{ij}=\Phi(z_{ij})\) and then applying the inverse empirical CDF, \(x_{ij}=\hat{F}_j^{-1}(u_{ij})\). More precisely, the inverse transformation selects the value \(x_{ij}\) corresponding to the interval of the empirical CDF that contains \(u_{ij}\).

For discrete variables, the CDF is a step function. Consequently, only the decoder, that is, the inverse transformation, is deterministic, since it represents a many-to-one mapping. The encoding operation, in contrast, is one-to-many: a single value in the data space can be represented by multiple latent states. To address this ambiguity, we adopt the concept of the distributional transform [cite] and assume that the latent variable is uniformly distributed within the admissible region. Here, we refer to the intermediate stage of the transformation, namely the quantile space. This concept will become clearer when we introduce the sampling procedure illustrated in Figure \ref{fig:orgf80f92a}. For the moment, however, let us focus on the derivation of the objective.
\subsection{Auxiliary objective function for discrete variables with continuous latent representation}
\label{sec:orgc7cc433}

The entropy term in the ELBO does not depend on \(\theta\), so for optimisation purposes we can focus on the joint distribution. Using the chain rule, for a given realisation \(z_i\), we have
\begin{equation}
p(X = x_i, Z = z_i \mid \theta) = p(X = x_i \mid Z = z_i, \theta)\, p(Z = z_i \mid \theta).
\end{equation}
The key consideration here is that we are modelling discrete data through the discretised response of a continuous latent. The conditional distribution \(p(X \mid Z, \theta)\) is degenerate, as it is induced by the empirical CDF responsible for the discretisation. For any \(z\) such that \(f(z) = x_i\), where \(f := \hat{F}^{-1} \circ \Phi\), the probability mass is equal to one, and zero elsewhere. To deal with this complex setting, we introduce the pre-image set
\begin{equation}
\mathcal{Z}_i = \{ z : x_i = f(z) \},
\end{equation}
and use it to define the limits of integration, which in turn facilitate the discretisation. The set \(\mathcal{Z}_i\) corresponds to the interval of latent values \(z\) for which the mapped quantile \(u = \Phi(z)\) lies within the bin associated with \(x_i\). This allows the problem to be expressed in terms of integrals over well-defined regions, thereby avoiding the discontinuities introduced by the empirical CDF.

Having introduced the pre-image sets \(\mathcal{Z}_i\), we can define the auxiliary function \(A(\theta)\), which we optimise in place of the ELBO. Since  \(x_i\) is associated with a set of admissible latent values rather than a single latent state, the objective integrates over the corresponding regions in the latent space forcing an introduction of observation-specific variational posterior \(q_i\), yielding
\begin{equation}
A(\theta) = \sum_{i} \int_{\mathcal{Z}_i} \log p(X=x_i, Z=z \mid \theta) \, q_i(z) \, dz.
\end{equation}
As noted earlier, the selection o \(q_i\) is, in principle, arbitrary. While in the standard ELBO formulation the optimal selection approaches the latent posterior, in our case, we have to deal with a whole set o \(q_i\), one per each region \(\mathcal Z_i\). We therefore introduce ansatz an observation-specific density obtained by restricting the single "parent" posterior to the corresponding pre-image and renormalising. We will write this as
\begin{equation}
q_i(z) = p(Z=z\mid X=x_i,\theta).
\end{equation}
While in this notation the result looks very standard, we have remember about the normalisation we discussed just a moment ago and that the condition on \(x_i\) has an important meaning here. In other words, each distinct point  \(x_i\)  redirect us to a pre-defined region of the latent space, where discretised response induces appropriate normalisation of the variational density.

As noted earlier, the choice of \(q_i\) is, in principle, arbitrary. In the standard ELBO formulation, the optimal choice approaches the latent posterior. In our case, however, we must deal with a family of distributions \(q_i\), one for each region \(\mathcal{Z}_i\). We therefore introduce an observation-specific density obtained by restricting a single "parent" posterior to the corresponding pre-image and subsequently renormalising it. We write
\begin{equation}
q_i(z) = p(Z=z\mid X=x_i,\theta),
\end{equation}
where \(p\) represent the truncated and normalised "parent". Although this expression appears entirely standard, it is important to keep in mind the normalisation discussed above and the specific role played by the conditioning on \(x_i\). Each observation \(x_i\) identifies a pre-defined region of the latent space, and the discretised response induces the corresponding normalisation of the variational density within that region. In other words, the particular/per-observation posterior is not defined over the entire latent space but only over the subset of latent states consistent with the observation under the empirical CDF mapping.

Finally, substituting the conditional structure obtained from the chain rule, we arrive at
\begin{equation}
\label{eq:orgb774c00}
\def\J{p(X=x_i \mid Z=z_i, \theta) p(Z=z_i \mid \theta)}
\def\P{p(Z=z_i \mid X=x_i, \theta)}
A(\theta) = \sum_{i} \int_{\mathcal{Z}_i} \log \left( \underbrace{\J}_{\text{joint distribution of data and latent var.}} \right)
\underbrace{\P}_{\text{probabilistic measure }} \, dz.
\end{equation}
At this point, the structure of the objective becomes transparent. The first factor in the joint distribution, \(p(X = x_i \mid Z = z, \theta)\), is degenerate: it is equal to one on the integration domain \(\mathcal{Z}_i\) and zero elsewhere. Since the integration is restricted to \(\mathcal{Z}_i\), this term remains constant throughout the integral and can therefore be omitted from further consideration. The second factor, \(p(Z = z \mid \theta)\), corresponds to the latent prior and contains the parameters of interest.

As a result, the objective takes the form of the expected log-likelihood of the latent prior over the admissible latent region associated with each observation.
\subsection{From a vicious cycle into a successful optimisation procedure.}
\label{sec:org45d290a}

The remaining difficulty lies in the conditional distribution \(p(Z = z \mid X = x_i, \theta)\), which determines how probability mass is distributed over the pre-image \(\mathcal{Z}_i\). Although all \(z \in \mathcal{Z}_i\) are consistent with the observation \(x_i\), they are not equally probable under the model. Their relative probability is governed by the latent prior \(p(Z \mid \theta)\), which is itself the object of estimation.

This creates a circular dependence that naturally leads to an expectation-maximisation (EM) formulation. Given a current parameter estimate \(\theta^{(k)}\), we first evaluate the posterior distribution over latent variables restricted to \(\mathcal{Z}_i\), namely \(p(Z \mid X = x_i, \theta^{(k)})\). This distribution assigns relative weights to the admissible latent states and constitutes the expectation step. We then update \(\theta\) by maximising the corresponding expected weighted log-likelihood under the current parameter estimate, yielding the maximisation step.
\subsection{Monte Carlo approximation and the self-normalising importance sampling (SNIS)}
\label{sec:orgc714397}

At this point, we need to address the computational challenges associated with the proposed objective. We have deliberately expressed it as a sum of expectations (see Equations \ref{eq:orgc1f24ce} and \ref{eq:orgb774c00}). The resulting difficulty is that we must evaluate many integrals, often in high-dimensional spaces. Fortunately, expectations of this kind can be approximated efficiently using Monte Carlo (MC) sampling:
\begin{equation}
\mathbb{E}_p[h(x)] \approx \frac{1}{N} \sum_{j} h(x_j),
\qquad \text{where } x_j \sim p(x),
\end{equation}
where, for the moment, \(h\) denotes an arbitrary well-behaved function. This approximation is particularly attractive in the context of probabilistic integrals.

We ultimately wish to sample from the distribution \(p(Z=z \mid X=x_i, \theta)\) restricted to the pre-image \(\mathcal{Z}_i\). However, given the large number of expectations that must be evaluated, even efficient sampling from such constrained regions can become challenging. In principle, one could resort to Markov chain methods or rejection sampling. In our case, however, both approaches are likely to be computationally prohibitive. For example, a straightforward implementation based on Markov chains would require maintaining a separate chain for every observation, each exploring its own admissible region in the latent space.

A better alternative is provided by importance sampling. Rather than drawing samples directly from each restricted distribution, we can sample from a more convenient proposal distribution and account for the discrepancy through appropriately chosen importance weights. As we will see, this allows the expectations appearing in the objective to be approximated quite efficiently.

Here, we arrive at one of the key elements of the proposed approach. For each observation \(x_i\), we draw samples uniformly from the corresponding quantile interval, in accordance with the distributional transform, and map them through the second stage of the transformation into the latent space. This procedure is illustrated in Figure \ref{fig:orgf80f92a}. In this way, we obtain a set of samples that provides good coverage of the associated regions \(\mathcal{Z}_i\). Although more importantly, the procedure guarantees that every observation contributes a controllable number samples from its admissible region in the latent space.

Moving forward, our ultimate goal, is to evaluate expectations with respect to \(p(Z = z \mid X = x_i, \theta)\), that is, the latent model restricted to the region \(\mathcal{Z}_i\). The samples constructed above do not follow this distribution. As a result, they cannot be used directly to estimate the expectation.

This is precisely where importance sampling becomes useful. Although the samples are generated from a more convenient proposal distribution, their contribution can be reweighted to recover expectations under the \(p(Z = z \mid X = x_i, \theta)\).

Consider a truncated distribution
\begin{equation}
p(x) = \frac{1}{C}\phi(x)\mathbb{I}_A(x),
\quad
C = \int_A \phi(x)\,dx,
\end{equation}
where \(C\) is the normalisation constant, \(\phi\) is the parent distribution, and \(A\) is the truncation volume. Note that these are general consideration and the notation is not related directly to the main derivation. The goal here is to calculate the expectation of a well-behaved function \(f\) under this distribution
\begin{equation}
\mathbb{E}_{p}[f(X)]
=
\int_V f(x) p(x) dx
=
\frac{1}{C} \int_V f(x)\phi(x)\,dx.
\end{equation}
Assume that we can only sample from \(X_i \sim q(x)\). The expectation can be expressed as a ratio of two integrals. The numerator is
\begin{equation}
\int_A f(x)\phi(x)\,dx
=
\int_A f(x)\frac{\phi(x)}{q(x)} q(x)\,dx
=
\mathbb{E}_q\!\left[f(X)\frac{\phi(X)}{q(X)}\right],
\end{equation}
and the denominator is
\begin{equation}
C
=
\int_A \phi(x)\,dx
=
\mathbb{E}_q\!\left[\frac{\phi(X)}{q(X)}\right].
\end{equation}
The result is:
\begin{equation}
\mathbb{E}_{p}[f(X)]
=
\frac{
\mathbb{E}_q\left[f(X)\frac{\phi(X)}{q(X)}\right]
}{
\mathbb{E}_q\left[\frac{\phi(X)}{q(X)}\right]
}.
\end{equation}
Given that \(X_1,\dots,X_N \sim q\),
\begin{equation}
\mathbb{E}_q[g(X)] \approx \frac{1}{N}\sum_{i=1}^N g(X_i).
\end{equation}
Then
\begin{equation}
\mathbb{E}_{p}[f(X)]
\approx
\frac{
\sum\limits_{i=1}^N f(X_i)\frac{\phi(X_i)}{q(X_i)}
}{
\sum\limits_{i=1}^N \frac{\phi(X_i)}{q(X_i)}
}.
\end{equation}
It immediately transpires that if
\begin{equation}
w_i := \frac{\phi(X_i)}{q(X_i)}.
\end{equation}
The above leads to
\begin{equation}
\mathbb{E}_{p}[f(X)]
\approx
\frac{\sum\limits_{i=1}^N w_i f(X_i)}{\sum\limits_{i=1}^N w_i}.
\end{equation}

This expression is exactly what we need to address the problem presented in this section. Our complicated sum over integrals can be reduced to a sum over weighted averages. 
\subsection{SNIS and the final version of the auxiliary objective in our EM framework}
\label{sec:orga67afaa}

Here, all components of the derivation come together. We have shown that Monte Carlo (MC) integration and self-normalized importance sampling (SNIS) can effectively address the challenges introduced by discrete variables in latent variable models. These challenges include evaluating the truncated form of the parent distribution, which must be normalized separately for each observation, and constructing a proposal distribution based on quantiles and data transformations. The latter, while provides a good coverage of the latent space, it introduces similar challenges.

The final step is to complete the definition of the EM auxiliary function and the corresponding M-step optimisation objective. This function now depends explicitly on both the current parameter estimate, \(\theta^{(k)}\), and the parameter vector being optimised in the current iteration, \(\theta\). For a single observation, in the k-th iteration, we approximate the integral
\begin{equation}
A_i \left( \theta, \theta^{(k)} \right) = \int_{\mathcal{Z}_i} \log p \left( X=x_i \mid Z=z, \theta \right) p \left( Z=z \mid \theta \right) \, p \left( Z=z \mid X=x_i, \theta^{(k)} \right) \, dz,
\end{equation}
where, as established earlier, the first term equals one throughout the integration region \(\mathcal{Z}_i\). Using \(M\) samples drawn from an observation-specific proposal distribution, we get
\begin{equation}
\def\W{\frac{p \left( Z=z_{ij} \mid X=x_i, \theta^{(k)} \right)}{\phi(z_{ij})} \frac{1}{\sum\limits_{j=1}^M \frac{p \left( Z=z_{ij} \mid X=x_i, \theta^{(k)} \right)}{\phi(z_{ij})}}}
A_i(\theta, \theta^{(k)}) \approx \sum_{j=1}^M \log p \left( Z=z_{ij} \mid \theta \right) \underbrace{\W}_{\text{normalised weight }} w_{ij}
\end{equation}
where \(z_{ij} \in \mathcal{Z}_i\)  are samples from the pre-image, \(\phi(z)\) is the distribution from the second stage of transformation -- the base for our proposal during the sampling that samples each feature independently. Here we change our notation slightly and the second index is not responsible for indexing the feature, but the sub-sample from approximating the probabilistic integral. This means we still operate with vectors of data and their latent representations.

Furthermore, \(p(Z=z_{ij} \mid \theta)\) now explicitely represents the current model and \(\theta\) the parameter vector that we want to optimise in the \(k+1\) step. The \(p(Z=z_{ij} \mid X=x_i)\) is the variational posterior. Here, thanks to application of SNIS we can replace it with the distribution that we refereed earlier as the "parent". In practice, this is our model under previous parameters \(\theta_k\).

Finally, the complete auxiliary objective is obtained by summing the contributions over all data points,
\begin{equation}
A(\theta, \theta^{(k)}) \approx \sum_{i=1}^N \sum_{j=1}^M w_{ij} \log p \left( Z=z_{ij} \mid \theta \right).
\end{equation}
The above result is essentially a well-known weighted likelihood problem which, for many models, including the Gaussian distribution, can be solved efficiently using maximum-likelihood estimates of \(\theta\) (the M-step). For example, in the Gaussian case, estimating the expectation reduces to computing a weighted average. The samples \(z_{ij}\), where \(j = 1, \ldots, M\), used to compute these estimates, form an "oversampled" representation of the latent state within the pre-image of \(x_i\), while the weights \(w_{ij}\) provide an SNIS 

In general, at the \$(k+1)\$-th maximisation step, the parameters \(\theta\) are estimated by maximising a weighted likelihood over the sampled latent representations. The weights are computed using the model obtained in the previous iteration and approximate the corresponding posterior probabilities. As the optimisation progresses, they are expected to become increasingly aligned with the target posterior distribution.

Since each observation \(x_i\) defines its own truncation region \(\mathcal{Z}_i\), the corresponding integral is approximated using an observation-specific proposal distribution. The importance weights \(w_{ij}\) must therefore be normalised separately for each observation, over its associated \(M\) latent samples. This is necessary because the proposal distribution depends on the truncation region and thus varies across data points. For the complete objective we just sum-up all contributions from all over-sampled data-points.

To summarise the key idea, each observation corresponds to a different discretisation of the continuous latent space and therefore to a different posterior distribution. These observation-specific posteriors are all induced by the same underlying latent model. By pooling evidence across these complementary "views" of the latent space, we can efficiently recover a single set of parameters \(\theta\) describing the underlying continuous distribution.

The complexity lies in finding the right latent representation. Once this is in place, the resulting algorithm is remarkably simple; see Section \ref{sec:orgba64f63} to see the description of the final optimisation loop.
\section{Deriving the ELBO and EM surrogate objective}
\label{sec:org4313ea1}
The Evidence Lower Bound (ELBO) provides one of the standard frameworks for dealing with latent-variable models. We introduce it by exposing the latent states and writing the likelihood of the observed data, given the model parameters, as a marginalised joint distribution
\begin{equation}
p(X=x \mid \theta)=\int p(X=x, Z=z \mid \theta)\,dz.
\end{equation}
Here, \(x\) represents a value in the data space, while \(z\) represents a latent state.

We now introduce an arbitrary variational density \(q(z)\) by multiplying and dividing the integrand by the same quantity. This allows us to rewrite the marginal likelihood as an expectation with respect to \(q\), i.e.
\begin{equation}
p(X=x \mid \theta)=\int q(z)\frac{p(X=x, Z=z \mid \theta)}{q(z)}\,dz
=\mathbb{E}_{q(z)}\left[\frac{p(X=x, Z=z \mid \theta)}{q(z)}\right].
\end{equation}

Since the same density appears in the numerator and denominator, this operation does not change the marginal likelihood. Taking the logarithm of both sides, as is standard in maximum-likelihood estimation, we express the log-likelihood as the logarithm of an expectation
\begin{equation}
\ell(\theta)=\ln\mathbb{E}_{q(z)}\left[\frac{p(X=x, Z=z \mid \theta)}{q(z)}\right].
\end{equation}

The problem is that it would be much more convenient to work with the expectation of a logarithm rather than with the logarithm of an expectation. We can achieve this by applying Jensen's inequality. In general, for a concave function \(f\)
\begin{equation}
f\left(\mathbb{E}[X]\right)\geq\mathbb{E}\left[f(X)\right].
\end{equation}
Since the logarithm is concave, the logarithm of an expectation is bounded from below by the expectation of the logarithm,
\begin{equation}
\ell(\theta)\geq\mathbb{E}_{q(z)}\left[\ln\frac{p(X=x, Z=z \mid \theta)}{q(z)}\right].
\end{equation}

We can now define the lower bound on the log-likelihood, which depends on the variational density and the model parameters. Expanding the logarithm gives the standard Evidence Lower Bound (ELBO)
\begin{equation}
\mathrm{ELBO}(q,\theta)
=\mathbb{E}_{q(z)}\left[\ln p(X=x, Z=z \mid \theta)\right]
-\mathbb{E}_{q(z)}\left[\ln q(z)\right],
\end{equation}
such that
\begin{equation}
\ell(\theta)\geq\mathrm{ELBO}(q,\theta).
\end{equation}
The ELBO consists of the expected joint log-likelihood and the entropy of the variational density,
\begin{equation}
H(q)=-\mathbb{E}_{q(z)}\left[\ln q(z)\right].
\end{equation}

When optimising with respect to \(\theta\) while keeping \(q\) fixed, the entropy term does not depend on the model parameters and can be omitted. The relevant objective is therefore the expected joint log-likelihood, commonly referred to as the EM auxiliary function or \$Q\$-function,
\begin{equation}
Q(\theta,q)=\mathbb{E}_{q(z)}\left[\ln p(X=x,Z=z \mid \theta)\right].
\end{equation}
\section{Choosing the variational posterior}
\label{sec:orgb690d9a}
The remaining question in Appendix \ref{sec:org4313ea1} was how to choose the variational density \(q\). To answer it, we consider the Kullback--Leibler divergence between \(q(z)\) and the true latent posterior \(p(Z=z\mid X=x,\theta)\):
\begin{equation}
D_{\mathrm{KL}}\!\left(q(z)\,\Vert\,p(Z=z\mid X=x,\theta)\right)
=\mathbb{E}_{q(z)}\left[\ln\frac{q(z)}{p(Z=z\mid X=x,\theta)}\right].
\end{equation}
Using Bayes' rule,
\begin{equation}
p(Z=z\mid X=x,\theta)=\frac{p(X=x,Z=z\mid\theta)}{p(X=x\mid\theta)},
\end{equation}
we can rewrite the KL divergence as
\begin{align}
D_{\mathrm{KL}}\!\left(q(z)\,\Vert\,p(Z=z\mid X=x,\theta)\right)
&=\mathbb{E}_{q(z)}[\ln q(z)] \\
&\quad-\mathbb{E}_{q(z)}[\ln p(X=x,Z=z\mid\theta)] \\
&\quad+\ln p(X=x\mid\theta).
\end{align}
The first two terms on the right-hand side form the negative ELBO. Rearranging therefore gives
\begin{equation}
\ln p(X=x\mid\theta)
=\mathrm{ELBO}(q,\theta)
+D_{\mathrm{KL}}\!\left(q(z)\,\Vert\,p(Z=z\mid X=x,\theta)\right).
\end{equation}

This decomposition of the log-likelihood provides a basis for interpreting the lower bound. Since the KL divergence is, by definition, non-negative,
\begin{equation}
D_{\mathrm{KL}}\!\left(q(z)\,\Vert\,p(Z=z\mid X=x,\theta)\right)\geq 0,
\end{equation}
the difference between the log-likelihood and the ELBO is always greater than or equal to zero. Furthermore, it is exactly the KL divergence between the variational density and the latent posterior:
\begin{equation}
\ln p(X=x\mid\theta)-\mathrm{ELBO}(q,\theta)
=D_{\mathrm{KL}}\!\left(q(z)\,\Vert\,p(Z=z\mid X=x,\theta)\right).
\end{equation}

For fixed model parameters \(\theta\), the bound is therefore maximised by choosing
\begin{equation}
q^\star(z)=p(Z=z\mid X=x,\theta).
\end{equation}
Under this choice, the KL divergence vanishes and the bound becomes tight:
\begin{equation}
\mathrm{ELBO}(q^\star,\theta)=\ln p(X=x\mid\theta).
\end{equation}

This result provides the basis for the expectation-maximisation procedure. For fixed model parameters, bringing \(q\) closer to the true posterior tightens the lower bound. Keeping \(q\) fixed and maximising the resulting bound with respect to \(\theta\), in turn, brings us closer to the maximum-likelihood estimate. The EM algorithm alternates between these two steps. This concept is illustrated in the figure
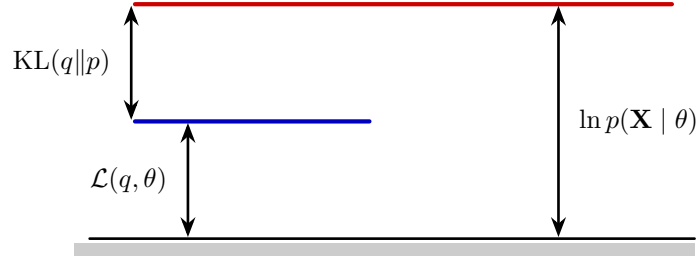
\begin{figure}[t]
\centering
\begin{tikzpicture}[line cap=round]
  \tikzset{
    base/.style   ={line width=1pt, black},
    topline/.style={line width=1.6pt, red!80!black},
    midline/.style={line width=1.6pt, blue!75!black},
    darr/.style   ={<->, line width=1pt, >=Stealth}
  }

  \def\xL{0}
  \def\xR{8.0}
  \def\yBase{0.0}
  \def\yMid{1.55}
  \def\yTop{3.10}

  \fill[gray!40] (\xL-0.2,\yBase-0.22) rectangle (\xR,\yBase-0.06);

  \draw[base] (\xL,\yBase) -- (\xR,\yBase);

  \draw[topline] (0.6,\yTop) -- (\xR-0.3,\yTop);

  \draw[midline] (0.6,\yMid) -- (3.7,\yMid);

  \draw[darr] (0.55,\yTop-0.02) -- (0.55,\yMid+0.02)
    node[midway, left=4pt] {$\mathrm{KL}\!\left(q\|p\right)$};

  \draw[darr] (1.30,\yMid-0.02) -- (1.30,\yBase+0.02)
    node[midway, left=4pt] {$\mathcal{L}(q,\theta)$};

  \draw[darr] (6.20,\yTop-0.02) -- (6.20,\yBase+0.02)
    node[midway, right=4pt] {$\ln p\!\left(\mathbf{X}\mid\theta\right)$};
\end{tikzpicture}
\caption{Relationship between the log-likelihood, the Evidence Lower Bound (ELBO), and the KL divergence. For fixed model parameters, bringing $q$ closer to the true latent posterior reduces the KL divergence and tightens the lower bound. Taken from \cite{Bishop2006}.}
\end{figure}
\section{Likelihood of a sample from the latent distribution (WIP)}
\label{sec:org0501fd4}
Results presented here are standard derivation that can be found in many textbooks. We include them here for convenience. 

Latent-variable models allow us to estimate the likelihood of an observation using the corresponding latent distribution. We assume a model
\begin{equation}
z\sim p(z):=p(z\mid\theta),
\end{equation}
together with an encoder and decoder,
\begin{equation}
z=f(x), \qquad x=g(z).
\end{equation}

Given a new observation \(x\), we wish to estimate its marginal likelihood \(p(x)\). From the probabilistic chain rule,
\begin{equation}
p(x)=\int p(x,z)\,dz=\int p(x\mid z)p(z)\,dz
=\mathbb{E}_{p(z)}[p(x\mid Z)].
\end{equation}
A direct Monte Carlo approximation can be obtained by drawing latent samples \(z_k\sim p(z)\) and averaging the corresponding conditional likelihoods as
\begin{equation}
\widehat{p}(x)=\frac{1}{K}\sum_{k=1}^{K}p(x\mid z_k).
\end{equation}
In practice, this estimate may be inefficient, since most samples drawn from the latent model can contribute very little to the likelihood of a particular observation. This is especially problematic when \(p(x\mid z)\) is concentrated within a small region of the latent space.

We can improve the estimate using importance sampling. Let \(q(z)\) be a proposal distribution chosen to place more samples in the region relevant to \(x\). Multiplying and dividing the integrand by \(q(z)\) gives
\begin{equation}
p(x)=\int\frac{p(x\mid z)p(z)}{q(z)}q(z)\,dz
=\mathbb{E}_{q(z)}\left[\frac{p(x\mid z)p(z)}{q(z)}\right].
\end{equation}
Given samples \(z_k\sim q(z)\), the likelihood can therefore be estimated as
\begin{equation}
\widehat{p}(x)=\frac{1}{K}\sum_{k=1}^{K}
\frac{p(x\mid z_k)p(z_k)}{q(z_k)}.
\end{equation}

The remaining question is how to choose the proposal distribution. Using Bayes' rule,
\begin{equation}
p(z\mid x)=\frac{p(x\mid z)p(z)}{p(x)},
\end{equation}
and rearranging,
\begin{equation}
p(x)=\frac{p(x\mid z)p(z)}{p(z\mid x)}.
\end{equation}
Comparing this expression with the importance-sampling estimate shows that the optimal proposal is the latent posterior,
\begin{equation}
q^\star(z)=p(z\mid x).
\end{equation}
Under this choice, the importance ratio is constant and equal to \(p(x)\), giving a zero-variance estimate. In practice, the exact posterior is usually unavailable, but this result provides a useful principle: the proposal should approximate \(p(z\mid x)\) as closely as possible.

We therefore use a conditional proposal \(q(z\mid x)\) and write
\begin{equation}
p(x)=\mathbb{E}_{q(z\mid x)}
\left[
\frac{p(x\mid z)p(z)}{q(z\mid x)}
\right]
\approx\frac{1}{K}\sum_{k=1}^{K}w_k,
\end{equation}
where
\begin{equation}
w_k=\frac{p(x\mid z_k)p(z_k)}{q(z_k\mid x)},
\qquad z_k\sim q(z\mid x).
\end{equation}
A simple example is
\begin{equation}
q(z\mid x)=\mathcal{N}\!\left(z\mid f(x),\sigma_p^2I\right),
\qquad
p(x\mid z)=\mathcal{N}\!\left(x\mid g(z),\sigma_o^2I\right),
\qquad
p(z)=\mathcal{N}(z\mid\mu,\Sigma).
\end{equation}
Here, \(q(z\mid x)\) places samples around the encoded representation of \(x\), \(p(x\mid z)\) measures how well the corresponding latent state reconstructs the observation, and \(p(z)\) evaluates that state under the fitted latent model.

For numerical stability, the calculation is performed in the logarithmic domain. We define
\begin{equation}
\ln w_k=\ln p(x\mid z_k)+\ln p(z_k)-\ln q(z_k\mid x).
\end{equation}
The log-likelihood estimate is then
\begin{equation}
\ln\widehat{p}(x)
=
\operatorname{LogSumExp}(\ln w_1,\ldots,\ln w_K)-\ln K.
\end{equation}

This provides a generic Monte Carlo estimate of the likelihood of an observation under a latent-variable model. Its precision depends primarily on how well the conditional proposal \(q(z\mid x)\) covers the parts of the latent space that make a substantial contribution to \(p(x)\).
\end{document}